\documentclass[runningheads]{llncs}
\usepackage[T1]{fontenc}
\usepackage{graphicx,verbatim}
\usepackage{booktabs,multicol,multirow}
\usepackage{amsmath}
\usepackage{bbding}
\usepackage{hyperref}
\usepackage{array}
\usepackage{caption}
\usepackage{subcaption}
\usepackage{url}
\usepackage{threeparttable}
\usepackage[T1]{fontenc} 
\usepackage{wrapfig}
\usepackage{tcolorbox}
\begin{document}
\title{IHC-LLMiner: Automated extraction of tumour immunohistochemical profiles from PubMed abstracts using large language models}
\titlerunning{IHC-LLMiner}
\author{Yunsoo Kim\inst{1}\orcidID{0009-0003-2244-5967} \and
Michal W. S. Ong\inst{2}\orcidID{0009-0001-7614-9018} \and
Daniel W. Rogalsky\inst{3}\orcidID{0000-0002-0194-8229} \and
Manuel Rodriguez-Justo\inst{2,4}\orcidID{0000-0001-5007-1761} \and
Honghan Wu\inst{1,5}\orcidID{0000-0002-0213-5668} \and
Adam P. Levine\inst{2,4}\orcidID{0000-0003-1333-9938}
}

\authorrunning{Kim et al.}

\institute{Institute of Health Informatics, University College London, 222 Euston Road, NW1 2DA, London, United Kingdom \and
Research Department of Pathology, University College London. University Street, WC1E 6DE, London, United Kingdom \and
Department of Pathology, Hadassah-Hebrew University Medical Center, Kiryat Hadassah, 12000, Jerusalem, Israel \and
Department of Cellular Pathology, University College London Hospitals NHS Foundation Trust, 60 Whitfield Street, W1T 4EU, London, United Kingdom \and
School of Health \& Wellbeing, University of Glasgow, 90 Byres Road, G12 8TB, Glasgow, United Kingdom\\
\email{\{yunsoo.kim.23,a.levine\}@ucl.ac.uk}}
    
\maketitle              
\begin{abstract}
Immunohistochemistry (IHC) is essential in diagnostic pathology and biomedical research, offering critical insights into protein expression and tumour biology. This study presents an automated pipeline, IHC-LLMiner, for extracting IHC-tumour profiles from PubMed abstracts, leveraging advanced biomedical text mining. There are two subtasks: abstract classification (include/exclude as relevant) and IHC-tumour profile extraction on relevant included abstracts. The best-performing model, ``Gemma-2 finetuned'', achieved 91.5\% accuracy and an F1 score of 91.4, outperforming GPT4-O by 9.5\% accuracy with 5.9 times faster inference time. From an initial dataset of 107,759 abstracts identified for 50 immunohistochemical markers, the classification task identified 30,481 relevant abstracts (\textit{Include}) using the Gemma-2 finetuned model. For IHC-tumour profile extraction, the Gemma-2 finetuned model achieved the best performance with 63.3\% \textit{Correct} outputs. Extracted IHC-tumour profiles (tumour types and markers) were normalised to Unified Medical Language System (UMLS) concepts to ensure consistency and facilitate IHC-tumour profile landscape analysis. The extracted IHC-tumour profiles demonstrated excellent concordance with available online summary data and provided considerable added value in terms of both missing IHC-tumour profiles and quantitative assessments. Our proposed LLM based pipeline provides a practical solution for large-scale IHC-tumour profile data mining, enhancing the accessibility and utility of such data for research and clinical applications as well as enabling the generation of quantitative and structured data to support cancer-specific knowledge base development. Models and training datasets are available at \url{https://github.com/knowlab/IHC-LLMiner}. 
\end{abstract}

\section{Introduction}
Immunohistochemical markers (IHC) are integral to daily histopathological practice, providing essential information for the diagnosis, prognostication, and treatment of various diseases, in particular, cancer \cite{duraiyan_applications_2012}. 
Although some IHC markers are supported by substantial evidence and knowledge regarding their sensitivity, specificity and clinical utility, others require further assessment in texts and online resources to evaluate their diagnostic and clinical relevance. The variability of IHC evidence in the literature and its disparate nature underscores the need for systematic approaches for IHC data evaluation and compilation \cite{fedchenko2014different,lin2014standardization,o2014garbage}.

A promising solution to streamline the collation, aggregation and analysis of IHC marker data is the application of text mining \cite{chang_automated_2013}. Text mining leverages computational techniques to extract and organise information from large volumes of unstructured textual data, such as the scientific literature and electronic health records (EHR). Early efforts utilising language models, particularly Bidirectional Encoder Representations from Transformers (BERT), for pathology report information extraction \cite{zeng_improving_2023} have shown encouraging results\cite{gao_biorex_2024}. These efforts demonstrate the feasibility of automating the identification of relevant markers and improving the efficiency of data extraction workflows.

Despite their potential, BERT models are primarily designed for discriminative tasks, such as document classification and named entity recognition (NER) \cite{masoumi_natural_2024,zeng_improving_2023}. This inherent design limitation poses challenges for their use in more complex text mining workflows. Specifically, to fully automate the extraction and alignment of IHC-related concepts, such as linking tumour types and tumour sites to their corresponding markers, BERT models would require training on two distinct tasks: NER and relation extraction \cite{masoumi_natural_2024}. This dual-task requirement significantly increases the complexity of preparing training datasets, as it demands extensive and detailed annotations. The annotation process, in turn, becomes more resource-intensive and expensive, limiting scalability and broad applicability.

Recent advances in large language models (LLMs), particularly following the introduction of ChatGPT and Gemini, represent a transformative step forward in biomedical text mining \cite{gemini2024capabilities,singhal2023towards,tu2023towards}. These models excel in generative and contextual language understanding tasks, often without the need for extensive task-specific training. Unlike traditional models, LLMs can be guided to produce desired outputs simply by providing example prompts, reducing the dependency on labour-intensive annotation processes. Their ability to link and synthesise information across vast textual datasets, including PubMed abstracts \cite{chataut_comparative_2024}, EHRs \cite{choi_developing_2023,johnson_large_2024}, and diagnostic pathology reports \cite{alzaid_large_2024,geevarghese_extraction_2025,truhn_extracting_2024}, opens up unprecedented opportunities. By enabling efficient extraction and alignment of data, these models significantly enhance the ability to uncover relationships and insights that were previously difficult to attain.

There have been limited efforts to aggregate IHC-tumour data from the literature. These have included both manually compiled commercial databases (e.g. ImmunoQuery \cite{noauthor_immunoquery_nodate}), online resources  (e.g. PathologyOutlines \cite{pathologyoutlinescom_stains_nodate} and Immunohistochemical Vade Mecum \cite{bishop2002immunohistochemical}) and reference texts (e.g. Quick Reference Handbook for Surgical Pathologists \cite{rekhtman2011quick}). However, these are either expensive, out-of-date, or unsystematic in nature. There is a need for an accurate, dynamic, comprehensive, freely-accessible database of IHC-tumour data. The objective of this study was to develop and implement a methodology using LLMs to systematically extract such data from PubMed abstracts and subsequently curate a structured and comprehensive database, facilitating histopathology diagnostic practice and research.

\section{Related Work}
\subsection{LLMs for Medicine}
ChatGPT greatly enhanced the exposure to and usability of LLMs through its interactive chat interface, allowing users to leverage such a model for a wide range of tasks, even those requiring intricate medical domain knowledge \cite{savage2024diagnostic}. However, proprietary LLMs, such as ChatGPT, face significant challenges when applied to large-scale text mining tasks in biomedical research. One of the key issues is the expense of using proprietary application programming interfaces (APIs), which can quickly escalate rendering them impractical for large-scale or resource-constrained projects. Furthermore, the reliance on external services introduces uncertainties in terms of cost fluctuations and long-term sustainability and affordability, further complicating their adoption for extensive biomedical applications.

Whilst proprietary models face limitations in terms of cost efficiency, open-source medical LLMs provide the advantage of local usage, even within an offline environment \cite{chen2023meditron,kim-2024-medexqa,toma2023clinical}. However, these open-source models come with their own set of challenges. Although they are more accessible, they still demand significant computational resources and, in many cases, underperform when it comes to understanding medical knowledge \cite{kim-2024-medexqa,longwell_performance_2024,wu_benchmarking_2024}. This creates a considerable barrier for many healthcare institutions, underscoring the need to develop more efficient models with improved performance, particularly for applications with in-depth biomedical knowledge.

\subsection{LLMs for Pathology}
In the field of pathology, LLMs are emerging as transformative tools for data extraction, diagnostic support and research standardisation. These models excel at processing unstructured or semi-structured data and have been tested for their ability to process histopathology reports \cite{alzaid_large_2024,geevarghese_extraction_2025,truhn_extracting_2024}. LLMs are able to extract key pathological elements, such as tumour histology and specimen type, with high accuracy. Despite their advantages, there remain challenges. LLMs sometimes struggle with ambiguous or poorly formatted inputs, requiring careful prompt design and expert review for validation.  

\section{IHC-LLMiner}

\begin{figure*}[!thbp]
    \centering
    \includegraphics[width=1\linewidth]{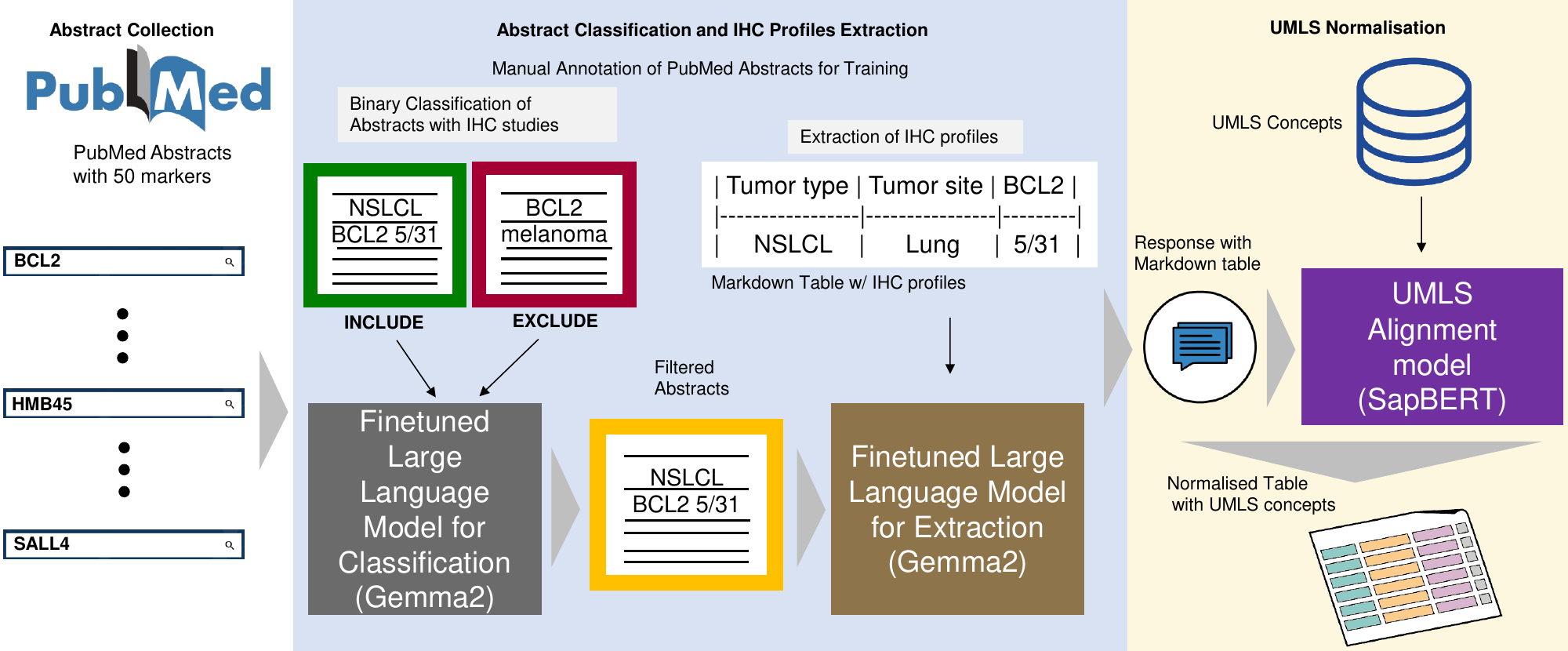}
    \caption{Overview of IHC-LLMiner: Automated extraction of Immunohistochemical profiles using LLMs.} 
    \label{fig:overview}
\end{figure*}

We developed IHC-LLMiner, an automated extraction pipeline composed of four key components: (i) PubMed Abstract Collection, (ii) Abstract Classification, (iii) IHC-Tumour Profile Extraction, and (iv) Unified Medical Language System (UMLS) Normalisation (Figure \ref{fig:overview}).

\subsection{PubMed Abstract Collection}

\begin{table}[!t]
\scriptsize
\caption{List of 50 Immunohistochemical Markers and Number of Downloaded Abstracts Per Marker\label{tab:markers}}
\begin{tabular*}{\columnwidth}{@{\extracolsep\fill}ll@{\extracolsep\fill}}
\toprule
\textbf{Marker} & \textbf{\# of Abstracts} \\
\midrule
ER               & 9,932  \\
p53              & 9,205  \\
CD3              & 8,187  \\
PR               & 8,095  \\
CD34             & 7,378  \\
BCL2             & 6,789  \\
HER2             & 6,230  \\
DESMIN           & 5,343  \\
p16              & 4,798  \\
S100             & 4,414  \\
SYNAPTOPHYSIN    & 4,105  \\
SMA              & 3,958  \\
CD10             & 2,355  \\
BRAF             & 2,285  \\
CALRETININ       & 2,179  \\
CHROMOGRANIN     & 1,983  \\
CD56             & 1,815  \\
p63              & 1,619  \\
AE1/AE3          & 1,573  \\
EMA              & 1,494  \\
CD20             & 1,405  \\
HMB45            & 1,227  \\
CD30             & 1,156  \\
CK7              & 992   \\
SOX10            & 947   \\
CA125            & 853   \\
SMAD4            & 731   \\
CD138            & 699   \\
WT1              & 680   \\
CDX2             & 593   \\
GATA3            & 543   \\
TTF1             & 530   \\
PAX8             & 434   \\
p40              & 426   \\
CD2              & 392   \\
CK20             & 340   \\
STAT6            & 298   \\
BCL6             & 245   \\
CK5              & 234   \\
BerEP4           & 224   \\
PLAP             & 200   \\
SALL4            & 193   \\
Brachyury        & 167   \\
DOG1             & 154   \\
BCL10            & 87    \\
MNF116           & 80    \\
BCL1             & 80    \\
MUC5             & 47    \\
CD168            & 40    \\
BOB1             & 25    \\
\hline
\end{tabular*}
\end{table}

The pipeline begins with the retrieval of abstracts from PubMed using Entrez e-utils \cite{kans2024entrez}. To ensure comprehensive coverage, search terms were formatted as \texttt{"MARKER immunohisto*"}, where \texttt{MARKER} represents the name of a specific IHC marker (e.g. BCL2, HMB45, SALL4). For this study, we selected 50 IHC markers commonly used in routine clinical diagnostic pathology practice. Due to limitations in Entrez e-utils, the maximum number of abstracts per marker was restricted to 9,999. Duplicate abstracts (arising from more than one IHC marker appearing in the same abstract) were identified and removed during preprocessing. In total, 107,759 abstracts remained for further analysis. The markers analysed in this study exhibit significant diversity in representation (Table~\ref{tab:markers}). Among the 50 markers, the most prevalent were ER and p53, with over 9,000 abstracts each. Other frequently reported markers, such as CD34, BCL2, and HER2, generated between 5,000 and 8,000 abstracts. Markers with moderate representation included Desmin, p16, and S100, with abstracts ranging from 1,000 to 4,999. In contrast, a substantial portion of markers (22 in total) had fewer than 500 abstracts, including BCL10, MNF116, and MUC5 with fewer than 100 abstracts.

\subsection{Abstract Classification}
To identify abstracts containing relevant IHC-tumour data, a binary classification approach was applied. A large language model that was fine-tuned with an expert annotations was employed. Abstracts classified as relevant were retained for downstream processing, whilst irrelevant abstracts were excluded from further analysis.

\subsubsection{Classification Data Annotation}
From the initial dataset of 107,759 abstracts, a random subset of 1,000 abstracts was manually annotated to train and evaluate the binary classification model. Each abstract was labelled as either \textit{"Include"} or \textit{"Exclude"} based on the following criteria:

\begin{itemize}
    \item The abstract reports the positivity rate of one or more IHC marker in one or more tumour types. Case reports describing IHC findings in a single patient are included.
    \item For multi-patient studies, the abstract must provide the exact number of patients who tested positive or negative for each marker in each tumour type.
    \item Review articles or meta-analyses were excluded to prevent data duplication.
\end{itemize}

\begin{table}[!t]
\caption{Classification Dataset Split\label{tab:classification_dataset}}%
\begin{tabular*}{\columnwidth}{@{\extracolsep\fill}llll@{\extracolsep\fill}}
\toprule
\textbf{Split} & \textbf{Size}  & \textbf{Include} & \textbf{Exclude} \\
\midrule
Train    & 800   & 364  & 436  \\
Evaluation    & 200   & 98  & 102  \\
Inference    &  106,759  & -  & -  \\
\hline
\end{tabular*}
\begin{tablenotes}%
\item Note: The inference dataset was not manually labelled but processed using the trained model.
\end{tablenotes}
\end{table}

This annotation process resulted in a dataset which was split into training and evaluation subsets, as shown in Table~\ref{tab:classification_dataset}.
The training subset contained 800 abstracts, with 364 labelled as \textit{Include} and 436 labelled as \textit{Exclude}. The evaluation subset contained 200 abstracts with 98 \textit{Include} and 102 \textit{Exclude}. Additionally, the remaining 106,759 abstracts were used as the inference dataset for the analysis of the 50 IHC markers.

The annotation process was conducted by three pathologists. A subset of annotations were cross-checked to ensure consistency and accuracy. This validation ensured that the classification model would allow only the most relevant abstracts to be included for downstream tasks, improving the reliability of the labelled dataset.

\subsubsection{Model Training and Evaluation} 
\begin{table}[!thbp]
\caption{Models for Abstract Classification\label{tab:classification_models}}%
\begin{tabular*}{\columnwidth}{@{\extracolsep\fill}llll@{\extracolsep\fill}}
\toprule
\textbf{Model} & \textbf{Parameters}  & \textbf{Domain} & \textbf{Type} \\
\midrule
BERT-base    & 110M   & General  & BERT  \\
DeBERTa-xsmall    & 70M   & General  & BERT  \\
Phi-3-mini    &  3.8B  & General  & LLM  \\
Gemma-2    &  9B  & General  & LLM  \\
GPT4-O    &  -  & General  & LLM  \\
\midrule
PathologyBERT    & 110M   & Clinical  & BERT  \\
SapBERT    &  110M  & Biomedical  & BERT  \\
BioLORD-2023    &  110M  & Biomedical  & BERT  \\
GatorTron    &  345M  & Biomedical  & BERT  \\
BiomedELECTRA-base    & 110M   & Biomedical  & BERT  \\
BiomedBERT-base    &  110M  & Biomedical  & BERT  \\
\hline
\end{tabular*}
\begin{tablenotes}%
\item Note: GPT4-O used here is the 2024-11-20 version.
\end{tablenotes}
\end{table}

We evaluated ten models, including five general-domain and five biomedical-domain models, to build a binary classification framework to identify relevant abstracts (Table~\ref{tab:classification_models}). We refer to models based on transformer encoder architecture as BERT-based models and models based on transformer decoder architecture as LLM. General-domain models consisted of three LLMs (Phi-3-mini, Gemma-2, GPT4-O) and two BERT-based models (BERT-base, DeBERTa-xsmall) \cite{abdin2024phi,devlin2018bert,he2021debertav3,hurst2024gpt,team2024gemma}. In contrast, at the time of the research, there were no publicly available biomedical versions of the Phi-3 and Gemma-2 models, so we only tested BERT-based biomedical models. These biomedical models included PathologyBERT (trained for pathology), SapBERT (fine-tuned on UMLS), BioLORD-2023 (fine-tuned on medical dictionary), GatorTron (trained on clinical notes), BiomedELECTRA-base, and BiomedBERT-base (both trained on PubMed abstracts) \cite{liu2020sapbert,remy-etal-2023-biolord,santos2023pathologybert,BioMedBERTandELECTRA,yang2022gatortron}.

The annotated dataset was used for training and evaluation. 
BERT-based models underwent standard fine-tuning by adding a classification head, while LLMs were aligned with the task via prompt engineering or task-specific fine-tuning with low-rank adaptation. Notably, GPT4-O was the only model evaluated without any additional fine-tuning.

Performance was assessed using classification accuracy and F1 score, with \textit{"Include"} as the positive class. Inference time and computational efficiency were also evaluated, particularly for larger models like Gemma-2 and GPT4-O, to assess their scalability for processing the full inference dataset.

\subsection{IHC-Tumour Profile Extraction}

For abstracts classified as relevant, IHC-tumour profiles were extracted and structured into markdown tables. These tables captured three key details: (i) tumour type, (ii) tumour site, and marker positivity rates (e.g., \texttt{5/31 for BCL2 in non-small cell lung cancer}). LLMs, such as Gemma-2, facilitated the accurate extraction and organisation of this information from unstructured text.

\subsubsection{IHC-Tumour Profile Data Annotation}

The 462 abstracts manually classified as \textit{"Include"} during the classification task were further annotated for IHC-tumour profiles extraction. Following the original train and evaluation split from the classification task, 364 abstracts were used for training and 98 for evaluation.

GPT4-O was used to generate initial markdown tables based on a structured prompt, which instructed the model to summarise tumour-specific and marker-specific findings into a tabular format. Annotators reviewed and corrected the tables for accuracy and completeness.

The prompt for GPT4-O included the following specific instructions for the task:
\begin{itemize}
    \item Listing tumour type and tumour site in separate columns.
    \item Reporting marker results as \texttt{X/Y}, where \texttt{X} represents positive cases and \texttt{Y} represents the total number of cases tested (\texttt{/1} for case reports).
    \item Avoiding grouping of tumour-marker results and marking unclear data as \texttt{NA}.
    \item Focusing solely on IHC findings without additional assumptions or extrapolations.
\end{itemize}

This prompt ensured standardised and accurate extraction of IHC data into a structured format suitable for normalisation.

\subsubsection{Human Evaluation}
To assess the performance of the fine-tuned model, pathologists performed a human evaluation on the entire evaluation split (98 abstracts). The model-generated markdown tables were compared to the annotated ground truth, with outputs categorised into three levels of accuracy:
\begin{itemize}
    \item \textbf{Correct}: The output perfectly matched the ground truth, with no errors or omissions.
    \item \textbf{Partially Correct}: The output contained some accurate information but included omissions or inaccuracies, reducing its practical/clinical utility.
    \item \textbf{Wrong}: The output significantly deviated from the ground truth, making it unreliable and unusable for pathologists.   
\end{itemize}

This evaluation provided valuable qualitative insights into the model's ability to extract the target IHC-tumour profile data. By involving pathologists, the evaluation ensured that the model's performance was assessed in terms of real-world usability, making it more relevant for downstream applications.

\subsubsection{LLM Extraction Fine-tuning with LoRA}
The best-performing open-source LLM was further fine-tuned for the IHC-tumour profile extraction task to enhance its performance. Given the limited size of the annotated dataset, we used Low-Rank Adaptation (LoRA) \cite{hu2021lora}, a parameter-efficient fine-tuning method. LoRA adapts the model for task-specific learning by introducing small, trainable rank-decomposition matrices while keeping the pre-trained model weights frozen. This approach is particularly advantageous for tasks with constrained data, as it mitigates overfitting and leverages the pre-trained knowledge of the LLM effectively.

LoRA enabled the efficient adaptation of the LLM to generate structured markdown tables summarising tumour types, tumour sites, and IHC marker results, optimising the model for precise extraction of IHC-tumour profiles from unstructured abstracts. Then, this fine-tuned model was applied to the inference dataset for the automated extraction of IHC-tumour profiles from unstructured abstracts, enabling accurate data generation at scale.

\subsection{Training and Evaluation settings}
For BERT-based models training, we used one A5000 GPU (24GB GPU) and the HuggingFace Transformers library \cite{wolf2019huggingface}. The training of BERT-based models used an AdamW optimizer with a learning rate of $5e^{-5}$ and a batch size of 4 for 2 epochs. For LLMs LoRA training, Phi-3 and Gemma-2, we used four A100 GPU (80GB GPU for each) and the LLaMA-Factory library \cite{zheng2024llamafactory}. We used a learning rate of $1e^{-4}$ for 3 epochs.

For evaluations, we used one A5000 GPU for all the models. For LLMs, we used a max new tokens of 4 for relevance classification and 1024 for the IHC profile extraction.

\subsection{UMLS Normalisation}
To ensure consistency and interoperability, the extracted IHC concepts were mapped to UMLS concepts \cite{UMLS_2024}. This mapping step utilised SapBERT \cite{liu2020sapbert}, a model specifically designed for aligning biomedical terms to UMLS. The final output was a normalised table with UMLS concepts facilitating downstream analysis.

To update SapBERT for use with the latest UMLS concepts, we fine-tuned the model using the UMLS2024AB metathesaurus release \cite{UMLS_2024}. This fine-tuning involved training the model for two epochs with a masked language model (MLM) objective to update its embeddings. The training data included concept names, trade names, and aliases, resulting in a dataset with 12,959,582 unique names. By including these variations, the model was optimised for recognising a broader range of biomedical terminology.

During inference, we used Euclidean distance within the embedding space to align extracted terms with their closest UMLS concepts. This approach ensured accurate and reliable normalisation of extracted IHC-tumour profiles to standardised biomedical concepts.

\subsection{IHC-Tumour Profile Landscape Analysis}
We conducted a landscape analysis of PubMed IHC-tumour data using IHC-LLMiner to evaluate the effectiveness of the automated workflow. The extracted data provided insights into tumour types, tumour sites and marker positivity rates. This automated analysis enabled the identification of patterns and trends in IHC-tumour data across PubMed abstracts.

To validate the pipeline, we compared the automated results of the 50 IHC markers and tumour pairs with known marker statistics from PathologyOutlines \cite{pathologyoutlinescom_stains_nodate}. For each IHC marker, we selected the top five tumour type based on the sample size. The largest sample sized tumour type with specific ranges or value for positivity rate is selected for comparison. If none of the markers have quantitative values, we use the explicit 'positive' or 'negative' label. In case, there are no qualitative values, we note the marker ``no data'' for the tumour type. These comparisons were used to assess the accuracy and reliability of the pipeline in extracting and structuring IHC-tumour associations. 

The analysis demonstrated the pipeline's potential to streamline research workflows, while also highlighted its utility in supporting large-scale literature reviews and systematic data aggregation for precision oncology.

\section{Results and Discussion}

\subsection{Abstract Classification}

\begin{table}[!b]
\caption{Performance Comparison of Models for Abstract Classification\label{tab:classification_perf}}%
\begin{tabular*}{\columnwidth}{@{\extracolsep\fill}llll@{\extracolsep\fill}}
\toprule
Model & Accuracy (\%)  & F1 & Run Time (s) \\
\midrule
Phi-3               & 63.5 & 61.0 & 28.3    \\
PathologyBERT       & 72.5 & 76.2 & 2.2     \\
DeBERTa-v3-xsmall   & 80.5 & 77.7 & 6.8     \\
BiomedBERT-base    & 84.0 & 80.7 & 2.1    \\
BiomedELECTRA-base     & 80.5 & 82.7 & \textbf{2.1}     \\
BioLORD-2023    & 82.5 & 83.1 & 6.9    \\
BERT-base           & 82.5 & 83.3 & 10.3     \\
GPT4-O             & 82.0 & 83.3 & 585.3  \\
SapBERT             & 84.5 & 85.2 & 5.6     \\
GatorTron           & 85.0 & 85.6 & 15.3    \\
Gemma-2             & \textbf{91.5} & \textbf{91.4} & 98.9    \\
\hline
\end{tabular*}
\begin{tablenotes}%
\item Note: Rows are sorted by F1 score. Run Time is the execution time for evaluating 200 abstracts. 
\end{tablenotes}
\end{table}

Various models were evaluated for their ability to classify PubMed abstracts for subsequent inclusion based on the aforementioned criteria.

Table~\ref{tab:classification_perf} presents the performance of individual models. Among the general-domain models, Gemma-2 demonstrated the highest accuracy (91.5\%) and F1 score (91.4). Its generative capabilities and large parameter size allowed it to outperform both smaller general models and some biomedical-specific BERT-based models.

Among biomedical-domain models, GatorTron achieved the highest performance (accuracy: 85.0\%, F1: 85.6) while maintaining computational efficiency with a runtime of 15.3 seconds for 200 abstracts. This result underscores the value of domain-specific pre-training. Despite being a smaller model, its alignment with biomedical text allowed it to achieve competitive results. Notably, both models outperformed GPT4-O (accuracy: 82.0\%, F1: 83.3), which, despite its advanced generative capabilities, struggled without fine-tuning the domain-specific classification task. GPT4-O also had inferior run time taking 5.9 times longer than Gemma-2 finetuned model.

The best-performing model, Gemma-2 fine-tuned model, was utilised for classifying abstracts in the inference dataset into \textit{Include} or \textit{Exclude} categories. This resulted in 30,481 abstracts classified as \textit{Include} out of 107,759 inference abstracts.

\subsection{IHC-Tumour Profile Extraction}
\begin{table}[!b]
\caption{Human Evaluation of IHC-Tumour Profile Extraction Model Outputs Against Ground Truth for 98 PubMed Abstracts. Rows are sorted by the Correct column. \label{tab:human_evaluation}}
\begin{tabular*}{\columnwidth}{@{\extracolsep\fill}lccc@{\extracolsep\fill}}
\toprule
\textbf{\multirow{2}{*}{Model}} & \textbf{\multirow{2}{*}{Correct (\%)}} & \textbf{Partially} & \textbf{\multirow{2}{*}{Wrong (\%)}}\\
 &   & \textbf{Correct (\%)} &  \\
\midrule
Gemma-2         & 2 (2.0\%)    & 12 (12.2\%)  & 84 (85.7\%) \\
Phi-3           & 13 (13.3\%)  & 29 (29.6\%)  & 56 (57.1\%) \\
Phi-3-FT        & 43 (43.9\%)  & 40 (40.8\%)  & 15 (15.3\%) \\
GPT4-O          & 58 (59.2\%)  & 38 (38.8\%)  & \textbf{2 (2.0\%)} \\
Gemma-2-FT      & \textbf{62 (63.3\%)}  & 28 (28.6\%)  & 8 (8.2\%) \\
\hline
\end{tabular*}
\end{table}

The task of extracting IHC-tumour profiles from the \textit{Include} abstracts involved summarising tumour type, tumour site and marker positivity rates into structured markdown tables. To evaluate the performance of different models, we conducted a human evaluation comparing outputs against the annotated ground truth (Table~\ref{tab:human_evaluation}).

The baseline Gemma-2 model, without any fine-tuning, performed the worst, with only 2.0\% of outputs classified as \textit{Correct}. It also produced the highest rate of \textit{Wrong} outputs (85.7\%), indicating that without domain adaptation, even large LLMs struggle with the precision required for biomedical information extraction. Phi-3, another foundation model, performed slightly better, achieving 13.3\% \textit{Correct} outputs and a lower error rate of 57.1\%, but remained inadequate for reliable extraction tasks.

Fine-tuning led to substantial performance improvements. Phi-3-FT achieved 43.9\% \textit{Correct} outputs and reduced \textit{Wrong} outputs to 15.3\%, demonstrating the effectiveness of task-specific adaptation. However, it still produced a large proportion of \textit{Partially Correct} outputs (40.8\%), suggesting that while overall fidelity improved, structural issues and partial omissions were still common.

\begin{figure*}[!b]
\centering
\begin{tcolorbox}[title=PMID: 21691200]
\scriptsize
... In this study, we evaluated the expression of S100A4 ... The study population included \textbf{155 clear cell renal cell carcinomas (cRCC)}, \textbf{22 papillary renal cell carcinomas (pRCC)}, \textbf{13 chromophobe renal cell carcinomas} and \textbf{13 oncocytomas}. ... S100A4 expression was significantly increased in the \textbf{stromal cells} in \textbf{cRCC (83\%)} and \textbf{pRCC (73\%)} ... There was no increased stromal cell expression of S100A4 in oncocytomas and chromophobe carcinomas. \textbf{Positive epithelial staining} was more common in \textbf{pRCC (58\%)} than \textbf{cRCC (11\%)}. ...
\end{tcolorbox}

\begin{tcolorbox}[title=Ground Truth, width=\textwidth]
\scriptsize
\begin{tabular}{|l|l|l|l|}
\hline
\textbf{Tumor type}                  & \textbf{Tumor site} & \textbf{S100A4 (epithelial)} & \textbf{S100A4 (stromal)} \\ \hline
Clear cell renal cell carcinoma & Kidney              & 17/155                     & 129/155                  \\ \hline
Papillary renal cell carcinoma  & Kidney              & 13/22                      & 16/22                    \\ \hline
Chromophobe renal cell carcinoma       & Kidney              & NA                         & 0/13                     \\ \hline
Oncocytoma                             & Kidney              & NA                         & 0/13                     \\ \hline
\end{tabular}
\end{tcolorbox}

\begin{tcolorbox}[title=Gemma-2: Incorrect, width=\textwidth]
\scriptsize
\begin{tabular}{|l|l|l|}
\hline
\textbf{Tumor type}                  & \textbf{Tumor site} & \textbf{IHC Marker 1} \\ \hline
Renal clear cell carcinoma & Kidney              & 83/155                       \\ \hline
Papillary renal cell carcinoma  & Kidney              & 73/22              \\ \hline
Papillary renal cell carcinoma  & Kidney              & 58/22              \\ \hline
Chromophobe renal cell carcinoma                             & Kidney              & NA/13             \\ \hline
Oncocytoma                             & Kidney              & NA/13                              \\ \hline
\end{tabular}
\end{tcolorbox}

\begin{tcolorbox}[title=GPT4-O: Partially Correct, width=\textwidth]
\scriptsize
\begin{tabular}{|l|l|l|}
\hline
\textbf{Tumor type}                  & \textbf{Tumor site} & \textbf{S100A4 (IHC)} \\ \hline
Clear cell renal cell carcinoma & Kidney              & 17/155                                      \\ \hline
Papillary renal cell carcinoma  & Kidney              & 13/22                                       \\ \hline
Chromophobe renal cell carcinoma       & Kidney              & 0/13                                        \\ \hline
Oncocytoma                             & Kidney              & 0/13                                        \\ \hline
\end{tabular}
\end{tcolorbox}

\begin{tcolorbox}[title=Gemma-2-FT: Correct, width=\textwidth]
\scriptsize
\begin{tabular}{|l|l|l|l|}
\hline
\textbf{Tumor type}                  & \textbf{Tumor site} & \textbf{S100A4 (Stromal)} & \textbf{S100A4 (Epithelial)} \\ \hline
Clear cell renal cell carcinoma & Kidney              & 128/155                   & 17/155                       \\ \hline
Papillary renal cell carcinoma  & Kidney              & 16/22                     & 13/22                        \\ \hline
Chromophobe renal cell carcinoma       & Kidney              & 0/13                      & NA                           \\ \hline
Oncocytoma                             & Kidney              & 0/13                      & NA                           \\ \hline
\end{tabular}
\end{tcolorbox}

\caption{Example of immunohistochemical-tumour profile extraction results from different models for PMID: 21691200.}
\label{fig:ihc_extraction}
\end{figure*}

Gemma-2 finetuned model was the best performing model, which reached 63.3\% \textit{Correct} outputs, outperforming GPT4-O by 4.1\%, and maintained a low \textit{Wrong} rate of 8.2\%. This demonstrates that fine-tuning Gemma-2 with LoRA offers a better contextual understanding than Phi-3 model architecture. Also, the result highlights the value of adapting LLMs to domain-specific extraction tasks making the model as a practical, scalable model for IHC–tumour profile extraction in biomedical contexts.

\subsection{IHC-Tumour Profiles Extraction Results}

Figure~\ref{fig:ihc_extraction} illustrates IHC-tumour profile extraction results across different models for a specific example. Among the models, Gemma-2-FT (fine-tuned) was the only one to produce a fully correct output. For stromal S100A4 expression in clear cell renal cell carcinoma (cRCC), Gemma-2-FT reported 128/155, compared to the ground truth of 129/155. This minor discrepancy is attributed to the abstract presenting the positive rate as a percentage (83\%), which is calculated to be 128.6 when applied to 155 cases. Both 129 and 128 were deemed correct due to this rounding.

GPT4-O achieved a \textit{Partially Correct} output by extracting epithelial expression data but omitting stromal expression results. Gemma-2 without fine-tuning, also produced a \textit{Partially Correct} result, reporting percentages normalised to 100 rather than actual case counts. Phi-3, another partially correct model, extracted stromal S100A4 expression for papillary renal cell carcinoma (pRCC) as 16/22 but failed to provide accurate stromal values for other tumour types, including cRCC, and misreported several key data points.

These findings highlight the clear benefits of fine-tuning Gemma-2 using LoRA. The fine-tuned model successfully captured both stromal and epithelial expression data with improved accuracy. Despite GPT4-O achieving strong overall performance, the gap between GPT4-O and Gemma-2-FT, as shown in this example, was not significant. Considering the modest performance gap between GPT4-O and Gemma-2-FT, along with the significantly higher cost of running GPT4-O for large-scale datasets, Gemma-2-FT emerges as a more practical choice for the inference dataset, balancing cost-effectiveness and extraction accuracy, particularly given the potential for ongoing prospective utilisation of this approach.

\subsection{IHC-Tumour Profile Landscape Analysis}

\begin{table*}[!b]
\caption{Top 10 Immunohistochemical Markers by Number of Abstracts\label{tab:ihc_markers}}
\tabcolsep=0pt
\begin{tabular*}{\textwidth}{@{\extracolsep{\fill}}lcccc@{\extracolsep{\fill}}}
\toprule
\textbf{IHC Marker} & \textbf{\# Abstracts} & \textbf{IHC Positives} & \textbf{Cohort Size} & \textbf{Positive Rate} \\
\midrule
VIMENTIN             & 4,177  & 21,104 & 31,263  & 67.5\% \\
S100                & 3,437  & 11,072  & 25,814  & 42.9\% \\
CD34                & 3,343  & 21,731 & 37,065  & 58.6\% \\
DESMIN               & 3,036  & 8,082 & 31,601  & 25.6\% \\
SMA                  & 2,951  & 14,070  & 32,282   & 43.6\% \\
EMA                  & 2,889  & 12,655 & 23,587  & 53.7\% \\
p53                  & 2,341  & 
53,442 & 111,423 & 48.0\% \\
CK7                & 2,270  & 20,700 & 34,953  & 59.2\% \\
SYNAPTOPHYSIN        & 2,102  & 10,721 & 22,618  & 47.4\% \\
CHROMOGRANIN                 & 2,026  & 14,375 & 31,723  & 45.3\% \\
\hline
\end{tabular*}
\end{table*}

The extracted IHC-tumour profiles from the inference dataset underwent post-processing to normalise terms and entities to UMLS concepts. This normalisation ensured consistency across marker names and tumour types, allowing for more robust and interpretable analyses. Using the UMLS-aligned data, we conducted a landscape analysis.

Table~\ref{tab:ihc_markers} presents the top ten IHC markers ranked by the total number of abstracts in which they are found. These markers represent a diverse set of proteins frequently evaluated and utilised in diagnostic pathology and research, reflecting their significance across tumour types and clinical contexts. The number of abstracts per IHC marker varied widely, with the most frequently studied marker, VIMENTIN, appearing in 4,177 abstracts, followed by S100, CD34, and DESMIN with over 3,000 abstracts each. The total cohort sizes reflected similar diversity. For instance, the cohort for p53 spanned over 110,000 cases, whereas markers such as DESMIN and SMA had cohort sizes closer to 30,000 cases. Positive rates varied from 25.6\% for DESMIN to 67.5\% for VIMENTIN.

\begin{table*}[!t]
\scriptsize
\caption{Comparison of IHC-LLMiner Extraction with PathologyOutlines (PO). When there is no data found in PathologyOutlines, we note it as ``\textbf{no data}.'' \label{tab:ihc_comparison}}
\begin{tabular*}{\textwidth}{@{\extracolsep{\fill}}lllll@{\extracolsep{\fill}}}
\toprule
\textbf{Marker} & \textbf{Tumour Type} & \textbf{IHC-LLMiner (\%)} & \textbf{PO (\%)} \\
\midrule
ER & luminal a breast cancer & 69 (2349/3384) & positive \\
p53 & infiltrating duct carcinoma &  31 (2638/8631) & \textbf{no data} \\
CD3 & mycosis fungoides & 98 (382/389) & positive \\
PR & infiltrating duct carcinoma & 50 (3225/6499) & 50-70 \\
CD34 & gastrointestinal stromal tumor & 61 (6712/11055) & 40-82 \\
BCL2 & diffuse large b-cell lymphoma & 65 (208/320) & 47-84 \\
HER2 & infiltrating duct carcinoma & 22 (6429/29087)   & 12-20 \\
DESMIN & gastrointestinal stromal tumor & 3 (258/7658) & negative \\
p16 & oropharyngeal squamous cell carcinoma & 58 (5345/9262) & \textbf{no data} \\
S100 & adenoid cystic carcinoma & 96 (97/101) & positive \\
SYNAPTOPHYSIN    & small cell carcinoma of lung & 78 (799/1024) &  positive \\
SMA              & leiomyosarcoma & 87 (409/471) & positive \\
CD10             & diffuse large b-cell lymphoma & 48 (882/1850) & 30-50 \\
BRAF             & papillary thyroid carcinoma & 79 (162/204) & positive \\
CALRETININ       & malignant mesothelioma & 88 (1356/1538) & 55-90 \\
CHROMOGRANIN     & carcinoid tumor & 83 (1955/2369) & positive \\
CD56             & small cell carcinoma of lung & 91 (1156/1264) & 90-100 \\
p63              & ovarian carcinoma & 63 (150/238) & \textbf{no data} \\
AE1/AE3          & pleural mesothelioma & 90 (133/147) & 84-100 \\
EMA              & meningioma & 72 (238/332) & positive \\
CD20             & classical hodgkin's lymphoma & 23 (155/677) & 20 \\
HMB45            & malignant melanoma of skin & 93 (757/816) & positive \\
CD30             & classical hodgkin's lymphoma & 97 (822/845) & positive \\
CK7              & chromophobe renal cell carcinoma & 91 (98/108) & positive \\
SOX10            & triple-negative breast carcinoma & 72 (604/841) & 58-74 \\
CA125            & endometrial carcinoma & 74 (442/599) & positive \\
SMAD4            & stomach carcinoma & 79 (400/504) & \textbf{no data} \\
CD138            & adenocarcinoma of colon & 40 (200/500) & \textbf{no data} \\
WT1              & high grade serous carcinoma & 93 (333/358) & 93 \\
CDX2             & stomach carcinoma & 36 (59/166) & 36-70 \\
GATA3            & luminal b breast cancer & 97 (2063/2118) & 91-100 \\
TTF1    & adenocarcinoma of lung            & 79 (3410/4315)      & 65–93 \\
PAX8             & renal cell carcinoma & 88 (447/509) & positive \\
p40              & squamous cell carcinoma & 95 (1088/1141) & positive \\
CD2              & lymphoma, extranodal nk-t-cell & 96 (156/162) & positive \\
CK20             & wolffian tumor & 0 (0/100) & negative \\
STAT6   & solitary fibrous tumor            & 97 (923/947)       & 98–100 \\
BCL6             & diffuse large b-cell lymphoma & 72 (189/264) & 60-90 \\
CK5              & squamous cell carcinoma & 99 (69/70) & 100 \\
BerEP4           & basal cell carcinoma & 58 (435/755) & 80-100 \\
PLAP             & retroperitoneal sarcoma & 77 (23/30) & \textbf{no data} \\
SALL4            & yolk sac tumor & 96 (188/195) & 100 \\
Brachyury        & chordoma & 87 (421/484) & 90 \\
DOG1             & colorectal carcinoma & 4 (65/1666) & 0-13 \\
BCL10            & diffuse large b-cell lymphoma & 31 (53/169) & \textbf{no data} \\
MNF116           & breast myoid hamartoma & 100 (24/24) & \textbf{no data} \\
BCL1             & myxoinflammatory fibroblastic sarcoma & 95 (138/146) & positive \\
MUC5             & sarcomatoid urothelial carcinoma & 23(6/26) & \textbf{no data} \\
CD168            & hormone sensitive prostate cancer & 31 (31/99) & \textbf{no data} \\
BOB1             & mediastinal (thymic) large b-cell lymphoma & 100 (139/139) & positive \\
\hline
\end{tabular*}
\end{table*}

To further evaluate the accuracy and clinical utility of the approach, the results for the 50 markers were compared with data from a widely accepted database for IHC markers. Table~\ref{tab:ihc_comparison} compares IHC marker positivity rates for major tumour types with those from PathologyOutlines \cite{pathologyoutlinescom_stains_nodate}. Among the 50 comparisons, 10 markers lacked corresponding data in PathologyOutlines, and 18 had only qualitative values (``positive'' or ``negative'') rather than specific ranges. All 18 markers with qualitative values had the IHC-LLMiner result aligning with the PathologyOutlines data. For the remaining 22 markers with quantitative values, 16 (72.7\%) fell within the reported ranges, while only one, BerEP4 in basal cell carcinoma, was notably outside the expected range.

These results reflect a strong concordance between the IHC-LLMiner extractions and an expert-curated source, underscoring the reliability of the system in extracting clinically meaningful IHC positivity rates. For example, IHC-LLMiner reported a TTF1 positivity rate of 79\% (3410/4315) in lung adenocarcinoma, closely matching the 65–93\% range in PathologyOutlines. Markers that deviated slightly from reference ranges may reflect variability in sample sizes, differences in antibody clones used, or heterogeneous data sources in the PubMed abstracts. For instance, STAT6 showed a 97\% positivity rate (923/947) in solitary fibrous tumour, just a slight difference with the reference range of 98–100\%.  This discrepancy (a 1\% offset) is likely within acceptable biological or experimental variability. 

Importantly, IHC-LLMiner also provides absolute case counts (e.g., 188/195 for SALL4 in yolk sac tumor), a feature not typically included in PatholoyOutlines. This enables more granular interpretation and direct quantitative comparisons, adding value to researchers and pathologists seeking high-resolution biomarker statistics.

Overall, the results highlight strong concordance between the IHC-LLMiner extractions and established pathology reference \cite{pathologyoutlinescom_stains_nodate}, underscoring the reliability of IHC-LLMiner for extracting accurate and clinically meaningful IHC marker profiles at scale. It supports the tool’s potential to augment expert resources, contribute to marker profile research across cancer subtypes, and enable automated knowledge base construction from biomedical literature.

\section{Conclusion}

In this study, we developed and evaluated an automated pipeline, IHC-LLMiner, for the classification and extraction of IHC-tumour profiles from PubMed abstracts. Leveraging both BERT-based models and LLMs, we achieved significant advancements in both classification accuracy and data extraction efficiency. 

For the classification task, Gemma-2 finetuned model achieved an impressive accuracy of 91.5\% and an F1 score of 91.4, outperforming GPT4-O, the state-of-the-art proprietary LLM. For IHC-tumour profile extraction, fine-tuning Gemma-2 with LoRA proved to perform the best with \textit{Correct} classifications at 63.3\%, higher than GPT4-O. Reliability and clinical utility of the IHC-LLMiner were also confirmed with the detailed analysis of selected 50 markers for the inference data, where the extracted positivity rates in all the markers except one aligned closely with established data available at PathologyOutlines. Moreover, 10 markers lacked corresponding data in PathologyOutlines, and 18 provided only qualitative labels such as "positive" or "negative," demonstrating IHC-LLMiner’s potential to fill gaps in existing pathology references with structured, quantitative data.

Despite these achievements, a key limitation of this study is the reliance on a relatively small annotated dataset for training and evaluating the fine-tuned Gemma-2 model for IHC-tumour profile extraction. Expanding the number of annotated samples could further enhance the model's ability to generalise across diverse biomedical texts and improve the quality of extracted data. Additionally, the normalisation step could benefit from integrating advanced LLMs, similar to the classification task, to further improve performance. Further optimised and improved models will permit the generation of an IHC-tumour profile database for clinical and research use.

Overall, IHC-LLMiner demonstrates the potential of integrating domain-specific and generative NLP models for automating complex text-mining tasks in biomedical research. By achieving a balance between cost-efficiency and performance, IHC-LLMiner offers a scalable and practical solution for large-scale IHC data analysis. Future work can address the limitations by increasing annotation efforts and extending the framework to additional biomedical datasets, further enhancing its applicability and impact.




\section{Conflict of interests}
No conflict of interest is declared.

\section{Author contributions statement}

APL conceived the study. YK designed and performed the computational analyses. MWSO, DR and APL generated the ground-truth annotations. YK and MWSO did the human evaluation. YK and MWSO drafted the manuscript with APL. APL, MRJ and HW provided supervision/guidance. All authors reviewed and approved the manuscript.

\section{Acknowledgments}
The authors kindly acknowledge funding from an NIHR Academic Clinical Lectureship (APL), The Jean Shanks Foundation/The Pathological Society of Great Britain \& Ireland (APL and MWSO) and Rosetrees Trust (APL and YK).
%
%
%
\bibliographystyle{splncs04}
\bibliography{main}
\end{document}